# Adversarial Sampling for Fairness Testing in Deep Neural Network

Tosin Ige[1], William Marfo[2], Justin Tonkinson[3], Sikiru Adewale[4], Bolanle Hafiz Matti[5]
Department of Computer Science, University of Texas at El Paso, Texas, USA[1, 2, 3]
Department of Computer Science, Virginia Tech. SW Blacksburg, Virginia. USA[4]
Department of Mathematics and Statistics, Austin Peay State University, Tennessee. USA[5]

***Abstract*—In this research, we focus on the usage of adversarial sampling to test for the fairness in the prediction of deep neural network model across different classes of image in a given dataset. While several framework had been proposed to ensure robustness of machine learning model against adversarial attack, some of which includes adversarial training algorithm. There is still the pitfall that adversarial training algorithm tends to cause disparity in accuracy and robustness among different group. Our research is aimed at using adversarial sampling to test for fairness in the prediction of deep neural network model across different classes or categories of image in a given dataset. We successfully demonstrated a new method of ensuring fairness across various group of input in deep neural network classifier. We trained our neural network model on the original image, and without training our model on the perturbed or attacked image. When we feed the adversarial samplings to our model, it was able to predict the original category/ class of the image the adversarial sample belongs to. We also introduced and used the separation of concern concept from software engineering whereby there is an additional standalone filter layer that filters perturbed image by heavily removing the noise or attack before automatically passing it to the network for classification, we were able to have accuracy of 93.3%. Cifar-10 dataset have ten categories of dataset, and so, in order to account for fairness, we applied our hypothesis across each categories of dataset and were able to get a consistent result and accuracy.**

## I. Introduction

With some of the latest advances in artificial intelligence, deep learning (DL) can now be applied in areas as diverse as, face recognition system [19], fraud detection system [20], and natural language processing (NLP) [21]. As deep neural network model continues to be increasingly associated with important decision in our daily life, we cannot just view it as only a mathematical abstraction but also as a technical system for the modern society[22],[18][17]. Apart from looking at the various metrics to better understand and evaluate the logic behind the prediction of machine learning model, it is also imperative to look at the ethics in order not to infringe on people's privacy in which the ability to ensure fairness across all groups without bias in dnn model prediction is of serious concern to the community[23], while it is possible to have intentional or unintentional discriminatory pattern in dataset[24] which are being used to train a dnn model, it is imperative and to have some mechanism to identify such discrimination in dataset before training model with it as such discrimination in dataset are eventually passed onto the trained model when the model is trained with discriminatory dataset especially when the discrimination is among the minority and the vulnerable in the society.

There are several forms in which discrimination can exists in dataset set, some of the forms includes group discrimination [25], [26] and individual discrimination [27], [28],[29]. Some of this discrimination can also be defined over a set of certain attributes such as gender, race age and so on., and when a machine learning model is trained with a discriminatory dataset, such discrimination in dataset is always passed to the trained model which make the model to make bias prediction over the same group being discriminated against in the dataset and this can be seen when ML model makes different decisions for different individuals (individual discrimination) or subgroups (group discrimination). Note that the set of protected attributes is often application-dependent and given in advance.

Our research work is focussed on the usage of adversarial sampling to test for the fairness in the prediction of deep neural network model across different classes of image in a given dataset. We are not dealing with the problem of individual discrimination or samples that differs only by some protected features. We aimed to use adversarial sampling to test for fairness in a dnn model, while also making an avenue for scaling the fairness through the misclassification rate across all group of image. Several adversarial samples were generated from the original image through several adversarial sampling techniques which includes Calini & Wagner, fast gradient sign method (FGSM), adversarial patch, gradient base evasion, and projected Gradient Descent (PGD). Although proposals and conceptual framework had been researched and formulated to address the issue of fairness in ML model [30], [31], [32]. One example is THEMIS randomly samples each attribute within its domain and identifies those discriminative samples [30], and also AEQUITAS which aims to improves the testing effectiveness with a two-phase (global and local) search [31] while SG [32] combines the local explanation of model [33] along with the symbolic execution [34] to cause an increment both in the discriminatory samples and diversity

## II. Background

Adversarial machine learning deals with the study of attacks on machine learning algorithms, and of the defenses against such attacks.[1] .Many years ago, the focus of machine learning engineers and scientist was on obtaining high accuracy for correct prediction, and while this had greatly been resolve in the past few years. The new challenge had focused on adversarial attack and defense against machine learning model there had been series of research survey which establish the need for protecting machine learning model against various forms of attach which make it to misclassify [2].

During the training of machine learning model, it is usually assumed that the training and test data are generated from the same statistical distribution. This assumption makes the final model vulnerable to various forms of attack, majority of which includes evasion attacks, [3] data poisoning attacks, [4] Byzantine attacks [5] and model extraction [6].

### A. Current Adversarial Techniques

*1) Gradient-based evasion attack:* In gradient base evasion attack, a perturbed image which seems like untampered to human eyes is made to be misclassified by neural network model (Fig. 1)[35].

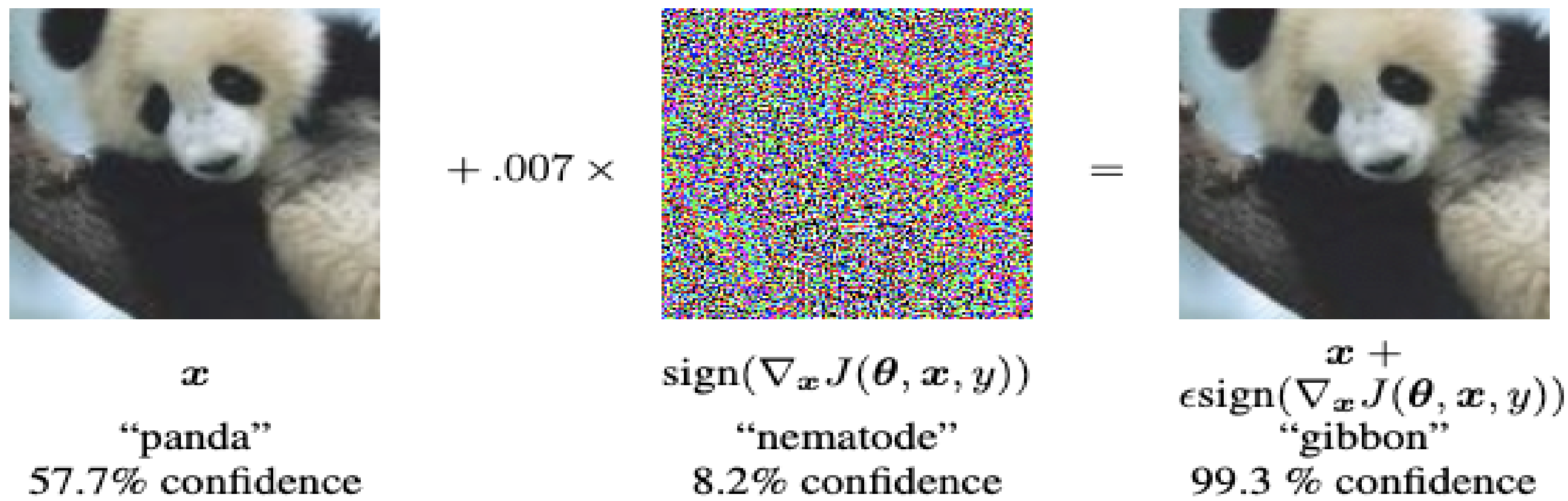

Fig. 1. Adversarial sampling based on addition of perturbation [35]

We can carry out this type of attack by trial and error method as we don't know in advance, the exact data manipulation that will break the model and make it to classify.

Let say we want to probe the boundaries of a machine learning model designed to filter out spam emails, it is possible for us to experiment by sending different emails to see what gets through. And so, a model has been trained for certain words like "momentum", and now we want to make an exceptions for emails that contains other words, if we want to attack, we can craft email with enough extraneous words which will eventually make the model to misclassify it.

### B. Fast Gradient Sign Method (FGSM)

Let's assume we want to produce an adversarial sample x' = x + η such that x' is misclassified by the neural network. For us to make x' and x produce different outputs, η should be greater than the precision of the features. Let's represent pixel of an image by 8 bits, and we want any information below 1/255 of the dynamic range to be discarded. Here, the input is perturbed with the gradient of the loss with respect to the input which gradually increases magnitude of the loss until the input is eventually misclassified.

While ε decides both the size and sign of each and every element of the perturbation vector which might be matrix or tensor which are being determined by the sign of the input gradient. Here, we just have to linearize the cost function and find the perturbation that maximizes the cost subject to an $L\infty$ constraint. This technique causes varieties of models to misclassify input and is also faster than other methods

*1) Projected gradient descent (PGD):* PGD initializes the sample to a random point in the ball of interest which is being decided by the $L\infty$ norm and does random restarts. This applies the same step as FGSM multiple times with a small step size while at the same time clipping the pixel values of intermediate results of each step to ensure that they are in an ε-neighborhood of the original image the value of α used is 1 which means that pixel values are changed only by 1 at each step while the number of iterations were heuristically chosen. This made it sufficient enough for the adversarial example to reach the edge of the ε max-norm ball.

*2) Carlini and wagner (C&W) attack:* Berkeley, Nicholas Carlini and David Wagner in 2016 propose a faster and more robust method to generate adversarial examples [7]. The attack proposed by Carlini and Wagner begins with trying to solve a difficult non-linear optimization equation. However instead of directly the above equation, Carlini and Wagner propose using a different function and then propose the use of the below function in place of f using z, a function that determines class probabilities for given input x.

With the use of stochastically gradient descent, we can use the above equation to produce a very strong adversarial sample especially when we compare it to fast gradient sign method which can effectively bypass a defensive distillation technique which was previously proposed for adversarial defense [7], [8], [9], [10].

*3) Adversarial patch attack:* Adversarial patch can be devised to fool a machine learning models. They work by causing physical obstruction in an image or by randomizing images with algorithm. Since computer vision models are trained on images that are straight forward. It is inevitable that any alteration to the input image can make the model to misclassify depending on the severity of the alteration.

We could define a patch function p corresponding to every transformation t € T which applies the transformed patch onto the image and Hadamard product, and the final adversarial perturbed image $\hat{x}$ which must satisfy $\hat{x} = p_t(x; \check{z})$ in order to trained patch $\check{z}$ and some t € T.

For us in order to train patch $\check{z}$, we could use a variant of the Expectation over Transformations (EOT) framework of Athalye et al. [3]. Let's assume a family of transformations T, a distance metric d in the transformed space, and the objective is to find a perturbed image $\hat{x}$

As the image is expected to be within €-ball in anticipation for transformations T. The attack could find some unconstrained optimization problem. The adversarial patch exploits the way machine learning model are trained for image classification by producing more salient inputs than real world objects. Such salient inputs are misclassified when fed to a machine learning model

### C. Current Defense Strategy and Limitation

*1) Adversarial training:* Adversarial training is a form of brute force supervised learning technique where several adversarial examples are fed into the model and are explicitly labeled as threatening. The approach is similar to a typical antivirus software, which is constantly being updated on a regular basis. As effective as adversarial training may be in defense against adversarial attack, it still requires continuous maintenance or update in order to be effective in combating new threats and it is still suffering from the fundamental problem of the fact that it can only successfully defend against threats or attack that has already happened and is already trained against.

*2) Randomization:* Several adversarial defense methods relied on randomization as a technique for mitigating the effects of adversarial Perturbations in the input and/or feature domain [11]. The idea behind this defense technique is the robustness of deep neural network model to random perturbation. The aim of randomization-based defense is to randomize the adversarial effects of the adversarial sampling into several random effects which is a very ok and normal thing for varieties of deep neural network models.

High success had been achieved by successful defense of Randomization-based defense technique against both black-box and gray-box based attacks, but it is still vulnerable white-box based attack, for example, the EoT method [12] can be easily attacked and compromised simply by considering the randomization process during attack.

*3) Denoising:* In denoising several research works had pointed to both input denoising and feature denoising as a technique for an effective adversarial defense. While input denoising is aimed at complete of partial removal of perturbation from the adversarial samplings or input, feature denoising is aimed at alleviating, reducing or mitigating effects of adversarial perturbation on important features i.e features that are more impactful on the decision of deep neural network model.

Several methods had been proposed for denoising as a technique for adversarial defense such as conventional input denoising, GAN-based input denoising, auto encoder-based input denoising, feature denoising.

Each of these methods of denoising had been shown to be vulnerable to one form of adversarial attack or another. For instance, Sharma and Chen [13] had shown that input squeezing can bypass by EAD, While good performance was achieved on Testbed by APE-GAN techniques[14], it is easily defeated by white-box based attack[16], As for auto encode-based input denoising, Carlini and Wagner [15],[16] successfully demonstrated that it is vulnerable to the adversarial samples generated by attack, but with feature denoising, research had shown that it merely increase accuracy by 3% which makes it vulnerable to PGD attack.

## III. Research Methodology

Fairness when using adversarial sampling as input had shown to cause disparity in accuracy and robustness among different groups [16]. Our method of approach is such as to investigate the cause and offer a solution. The methodologies were in three(s) phases of activities;

### A. Phase-1: Development and Optimization of DNN Model from Scratch

We created a python project in jupyter notebook and created a deep neural network model (DNN) for image classification. We used cifar-10 dataset which consists of 60000 colored images with each image having 32x32 dimensions and categorized into 10 classes of image [Airplane, Automobile, Bird, Cat, Deer, Dog, Frog, Horse, Ship, Truck], with each category containing 6000 colored images. The whole 50000 images in the training dataset folder of cifar-10 was used for training our DNN model while the 10000 images in the test folder of cifar-10 dataset was used to validate our model.

With an initial accuracy of 72%, we needed a higher accuracy and hence, we did some hyper parameter turning by adjusting the learning rate, the number of convolusional layers, and adding some regularization until we are able to get a good accuracy that we can work with, after which we decide to compile and save our new model with KERAS being a high level neural network library that runs on tensorflow.

### B. Phase-2: Generation of Adversarial Sampling

Another python class was created in jupyter notebook in which we wrote algorithms for several adversarial attacks. We wrote adversarial algorithm for adding various kind of noise perturbation to all the images in the training folder (adversarial sampling) (Fig. 2). The algorithm automatically creates new folder and then puts all the adversarial sampling into the new folder. The adversarial folder which contains all the perturbed image or adversarial sampling is named **dogsa** (Fig. 3).

```
def noise_add(path,numb):
    img = cv2.imread(path,0)
    im = np.zeros(img.shape, np.uint8)
    mean=40
    sigma=50
    cv2.randn(im,mean,sigma) # create the random distribution
    noise_image = cv2.add(img, im) # add the noise to the original image
    io.imsave("dogsa\\"+str(numb)+".jpg", noise_image)
    return noise_image
```

Fig. 2. Creation and addition of random gaussian noise distribution to image having image path and position as argument

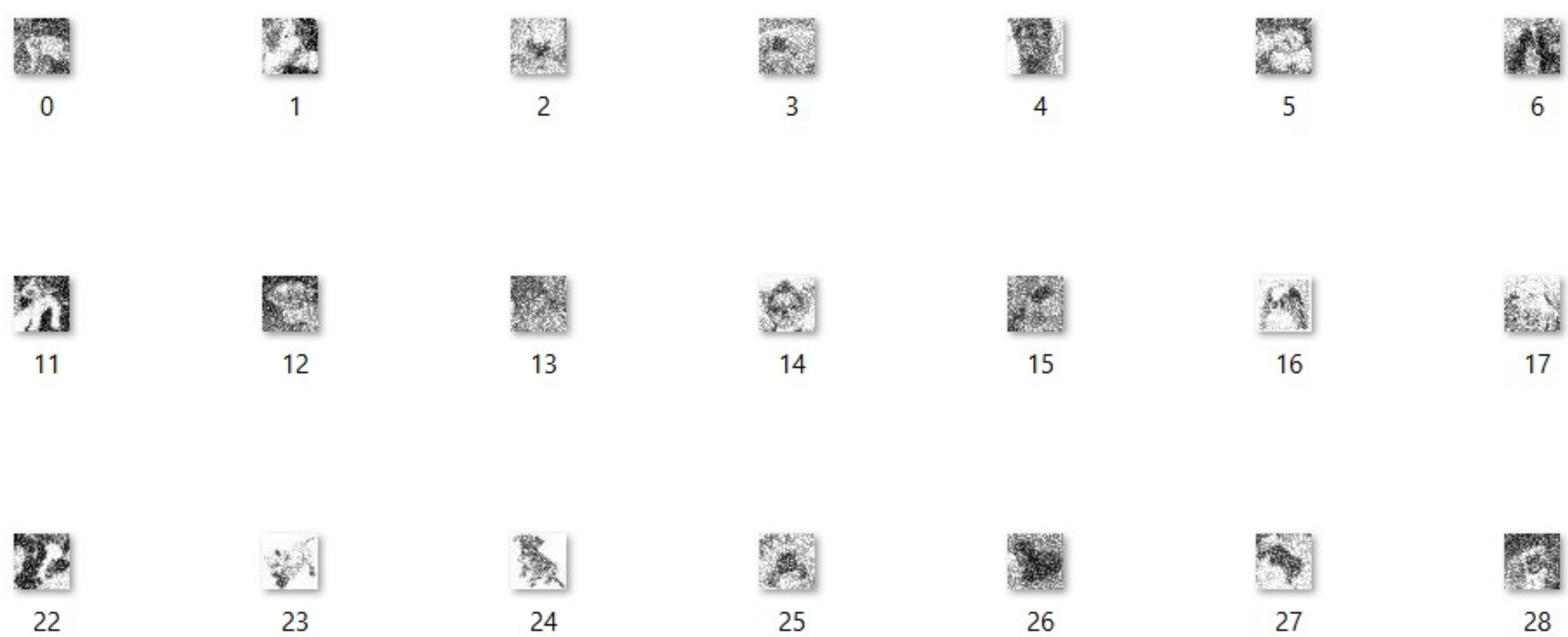

Fig. 3. Generated adversarial sampling in the dogsa folder after iteration and noise attack

At this point, it was needful for us to test our model with the newly created adversarial sampling to see if it will misclassify those images, having satisfied the criteria of misclassification, it was needful for us to test for fairness across each of the 10 categories of images in the cifar-10 dataset. To actualize this, we separated each category of image in the cifar-10 dataset as separate entity and then observe the accuracy of misclassification across each entity to see if the accuracy of misclassification for each entity will be close, and indeed the accuracy of misclassification for each of the 10 categories of images were closely called which ensure fairness across each group.

## C. Phase-3: Evaluation and Removal of perturbation

This is a very tricky part, as several methods had been proposed with little or no effectiveness. Here, we write algorithm to remove the perturbation, considering the existence of several adversarial attack, we wrote an algorithm to remove all forms of perturbation while at the same trying to maintain the original property of the image. The algorithm iterate through all the images in the adversarial folder where we have our adversarial samplings and remove perturbation in each of them, and in the process creating a new folder name **dogsa-clean** where all the clean images from the adversarial folder are saved (Fig. 4).

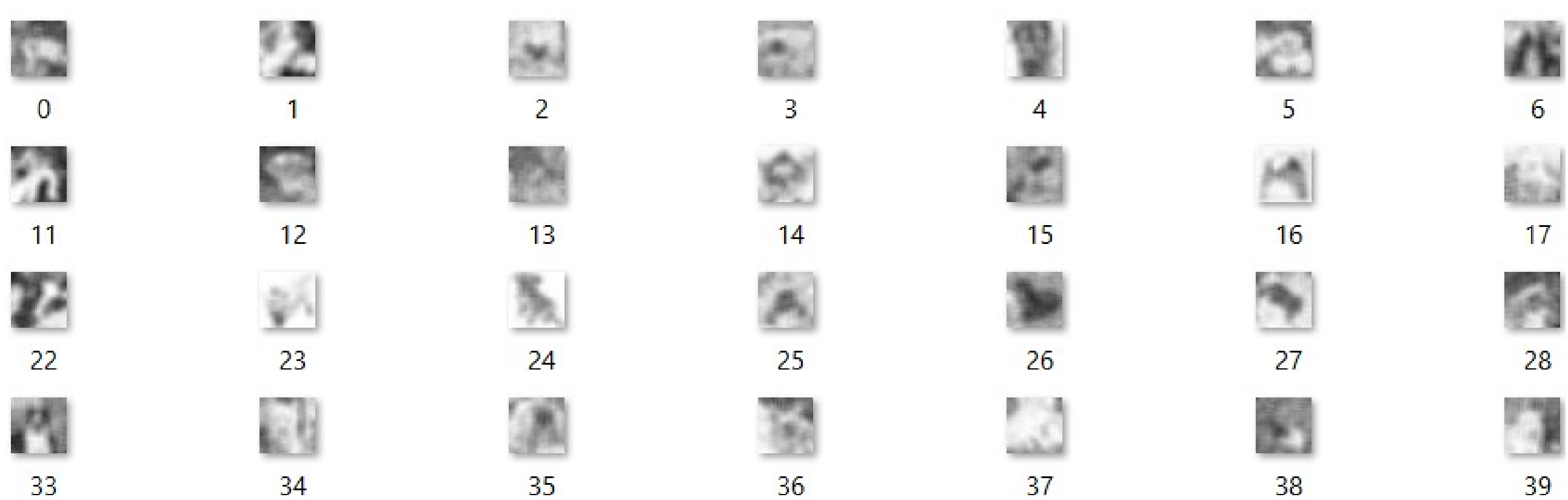

Fig. 4. Images in the dogsa-clean folder after passing through the new separation of concern layer from software engineering concept

At this point, we wrote few lines of python code to load our model through keras, and iterating through each of the images in the **dogsa-clean** folder where the cleaned images are saved and observe the result. After this, each categories of image were treated as separate entity to account for fairness across each group of images.

## IV. Result and Discussion

On iterating through each category of adversarial samplings in our adversarial folder to see the rate of misclassification across each image, it was found that there is unfairness as some of classes have high rate of positive misclassification than the others.

The rate of misclassification was not consistent as Airplane, Automobile, and truck (Fig. 5) has very low rate of misclassification compared with other groups, calini & wagner form of adversarial attack were added to them, while also updating the learning rate and regularization of our initial model and rebuild. The purpose of this is to ensure fairness and consistency across all the classes of image through consistence rate of misclassification (Fig. 6).

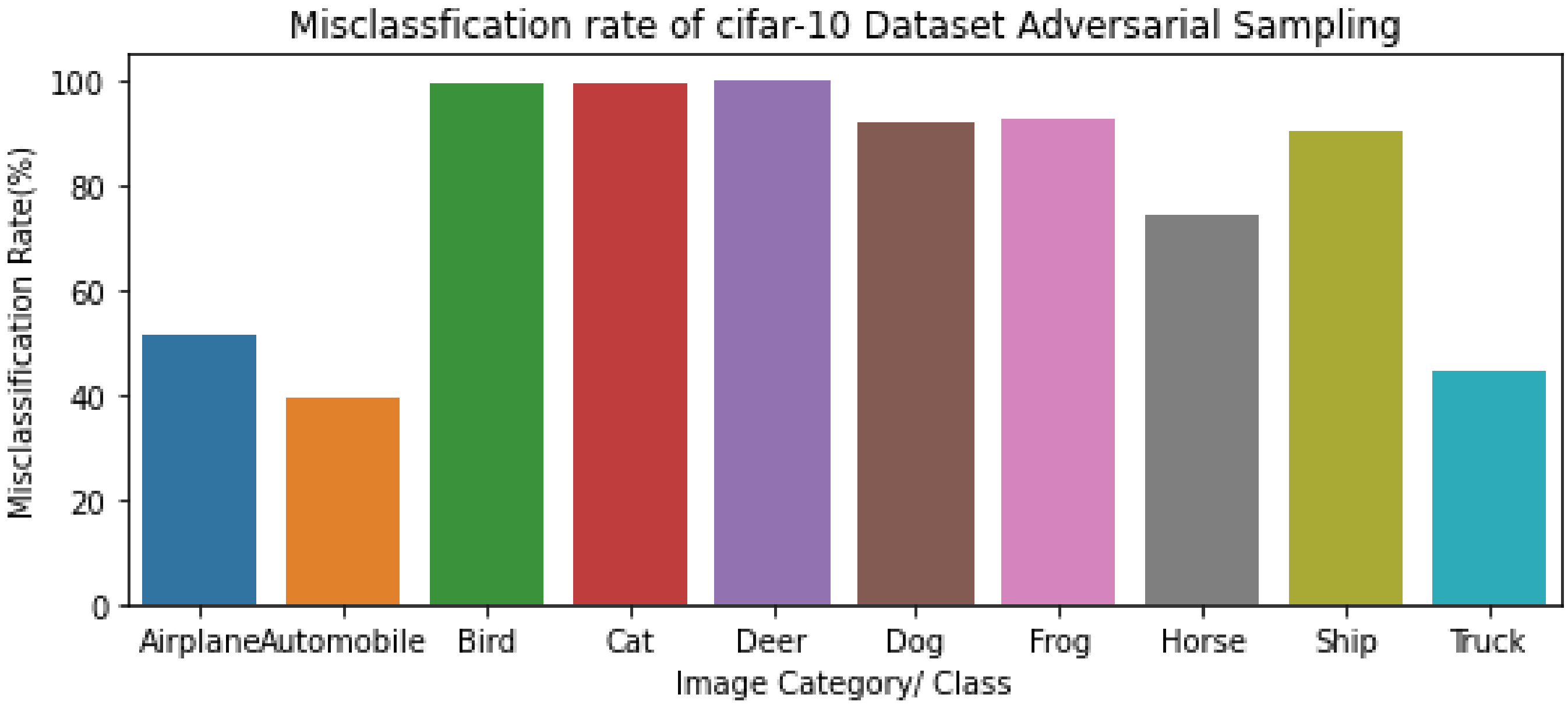

Fig. 5. Plotting of the misclassification rate of the generated adversarial sampling and visualization of fairness across each of the 10 category of images

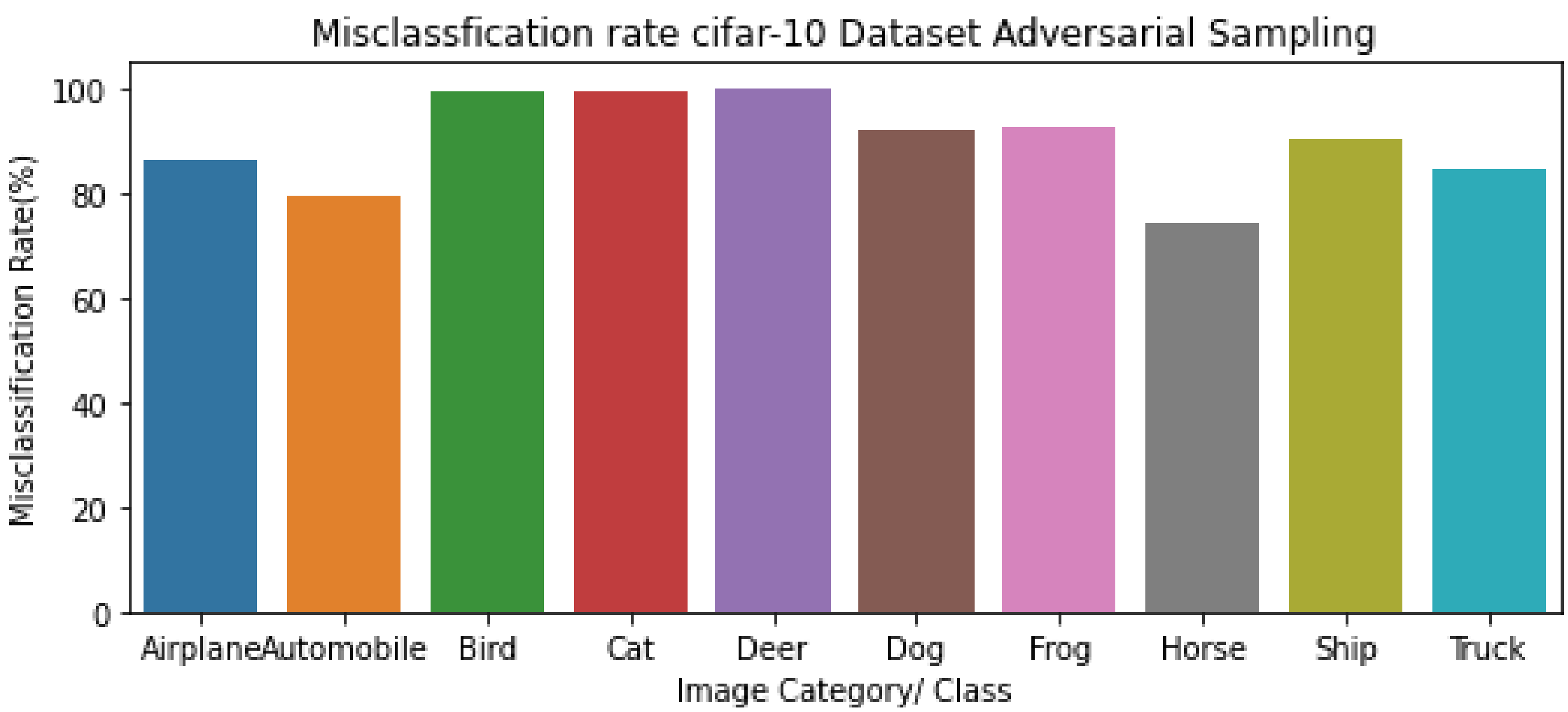

Fig. 6. Plotting of the misclassification rate of the generated adversarial sampling and visualization of fairness across each of the 10 category of images after Calini & wagner attack

With some consistencies in the rate of misclassification, series of algorithms were written to remove greater part of the noises and perturbation for better accuracy. Rather than labeling the attacked images as invalid input, we want the model to be able to predict the attacked images correctly and classify them to the category of images they belonged to.

We wrote an algorithm to iterate through all the images in our adversarial folder, remove as much pertubation and denoise as possible while creating a new folder **dogsa-clean** to store all the new images. Substantial lines of python codes where written to iterate through various classes of cleaned and denoised images in our dogsa-clean folder, and feeding them to the DNN model.

To our surprise and amazement, without making any further hyper-parameter tuning, we were able to have high rate of fairness and consistency across each of the classes of clean images while still maintaining a very good accuracy of prediction in our classifier.

## V. Conclusion

In this research, we successfully demonstrate a new method of ensuring fairness across various group of input in deep neural network classifier. Rather than the existing method of training the model on the adversarial sample and label them as invalid. We trained our neural network model on the original image, and without training our model on the perturbed or attacked image. When we feed the adversarial samplings to our model, it was able to predict the original category/ class of the image the adversarial sample belong to.

Through our, method we were able to achieve fairness across all the categories of images in the cifar-10 dataset. We also introduce Separation of Concern (SOC) method from full stark software engineering which ensures that we can manage the filter layer as separate entity at any point in the development life cycle without re-training the model.

Surprisingly, we tried to compare the true rate and false rate of fairness for the adversarial sampling across each of the classes of image with that of the cleaned images. We found that the fairness rate was high, consistent and almost the same without any hyper parameter tuning or modification to the filter layer.

## VI. Limitation and Future Research

In this research, we use the existing forms of adversarial attack for the images. However, we envisage that there will be more sophisticated forms of attack in the future. It is for this reason that we adopt the model of separation of concern from software engineering for our filter layer. In the event of a more robust and sophisticated attack, rather than going back to development to retrain our model, we only need to improve the filter layer.

In addition, the filter layer can be made into a cloud based handy toolbox library. In that case, the filter layer can be managed in the cloud against any future robust attack and be automatically available to all existing deployed model.

### A. Material and Source

We use python 3.9 for this project, pip version 22.3.1, tensorflow, keras to save, load and consume our model and a host of other python libraries.

Our source code for the hypothesis and experiment on this research had been uploaded to github and is made available to the public, and can be accessed through the Uniform Resource Locator (URL) below: https://github.com/IGETOSIN1/Research-Adversarial-Sampling-for-Fairness-Testing

## References

[1] Kianpour, Mazaher; Wen, Shao-Fang (2020). "Timing Attacks on Machine Learning: State of the Art". Intelligent Systems and Applications. Advances in Intelligent Systems and Computing. Vol. 1037. pp. 111–125. doi:10.1007/978-3-030-29516-5_10. ISBN 978-3-030-29515-8. S2CID 201705926.

[2] Jump up to:a b Siva Kumar, Ram Shankar; Nyström, Magnus; Lambert, John; Marshall, Andrew; Goertzel, Mario; Comissoneru, Andi; Swann, Matt; Xia, Sharon (May 2020). "Adversarial Machine Learning-Industry Perspectives". 2020 IEEE Security and Privacy Workshops (SPW): 69–75. doi:10.1109/SPW50608.2020.00028. ISBN 978-1-7281-9346-5. S2CID 229357721.

[3] Goodfellow, Ian; McDaniel, Patrick; Papernot, Nicolas (25 June 2018). "Making machine learning robust against adversarial inputs". Communications of the ACM. 61 (7): 56–66. doi:10.1145/3134599. ISSN 0001-0782. Retrieved 2018-12-13.[permanent dead link]

[4] Geiping, Jonas; Fowl, Liam H.; Huang, W. Ronny; Czaja, Wojciech; Taylor, Gavin; Moeller, Michael; Goldstein, Tom (2020-09-28). Witches' Brew: Industrial Scale Data Poisoning via Gradient Matching. International Conference on Learning Representations 2021 (Poster).

[5] Jump up to:a b c El-Mhamdi, El Mahdi; Farhadkhani, Sadegh; Guerraoui, Rachid; Guirguis, Arsany; Hoang, Lê-Nguyên; Rouault, Sébastien (2021-12-06). "Collaborative Learning in the Jungle (Decentralized, Byzantine, Heterogeneous, Asynchronous and Nonconvex Learning)". Advances in Neural Information Processing Systems. 34. arXiv:2008.00742.

[6] Tramèr, Florian; Zhang, Fan; Juels, Ari; Reiter, Michael K.; Ristenpart, Thomas (2016). Stealing Machine Learning Models via Prediction {APIs}. 25th USENIX Security Symposium. pp. 601–618. ISBN 978-1-931971-32-4.

[7] Carlini, Nicholas; Wagner, David (2017-03-22). "Towards Evaluating the Robustness of Neural Networks". arXiv:1608.04644 [cs.CR].

[8] "carlini wagner attack". richardjordan.com. Retrieved 2021-10-23.

[9] Plotz, Mike (2018-11-26). "Paper Summary: Adversarial Examples Are Not Easily Detected: Bypassing Ten Detection Methods". Medium. Retrieved 2021-10-23.

[10] Wang, Xinran; Xiang, Yu; Gao, Jun; Ding, Jie (2020-09-13). "Information Laundering for Model Privacy". arXiv:2009.06112 [cs.CR].

[11] Kui Ren, Tianhang Zheng, Zhan Qin, Xue Liu, Adversarial Attacks and Defenses in Deep Learning, Engineering,Volume 6, Issue 3,2020,Pages 346-360,ISSN 2095-8099, https://doi.org/10.1016/j.eng.2019.12.012

[12] Athalye A, Engstrom L, Ilya A, Kwok K. Synthesizing robust adversarial examples. 2017. arXiv:1707.07397.

[13] Sharma Y, Chen PY. Bypassing feature squeezing by increasing adversary strength. 2018. arXiv:1803.09868.

[14] Shen S, Jin G, Gao K, Zhang Y. APE-GAN: adversarial perturbation elimination with GAN. 2017. arXiv: 1707.05474.

[15] Rokach L. Decision forest: twenty years of research. Inf Fusion. 2016;27:111–25.

[16] https://doi.org/10.48550/arxiv.2010.06121, doi = {10.48550/ARXIV.2010.06121}, url = {https://arxiv.org/abs/2010.06121}, author = {Xu, Han and Liu, Xiaorui and Li, Yaxin and Jain, Anil K. and Tang, Jiliang}, keywords = {Machine Learning (cs.LG), Machine Learning (stat.ML), FOS:

Computer and information sciences, FOS: Computer and information sciences}, title = {To be Robust or to be Fair: Towards Fairness in Adversarial Training}, publisher = {arXiv},year = {2020}, copyright = {arXiv.org perpetual, non-exclusive license}}

[17] P. Zhang et al., "Automatic Fairness Testing of Neural Classifiers Through Adversarial Sampling," in IEEE Transactions on Software Engineering, vol. 48, no. 9, pp. 3593-3612, 1 Sept. 2022, doi: 10.1109/TSE.2021.3101478.

[18] P. Zhang et al., "Automatic Fairness Testing of Neural Classifiers Through Adversarial Sampling," in IEEE Transactions on Software Engineering, vol. 48, no. 9, pp. 3593-3612, 1 Sept. 2022, doi: 10.1109/TSE.2021.3101478.

[19] F. Schroff, D. Kalenichenko and J. Philbin, "Facenet: A unified embedding for face recognition and clustering", Proc. IEEE Conf. Comput. Vis. Pattern Recognit., pp. 815-823, 2015.

[20] K. Fu, D. Cheng, Y. Tu and L. Zhang, "Credit card fraud detection using convolutional neural networks", Proc. 23rd Int. Conf. Neural Inf., pp. 483-490, 2016.

[21] E. Wulczyn, N. Thain and L. Dixon, "Ex machina: Personal attacks seen at scale", Proc. 26th Int. Conf. World Wide Web, pp. 1391-1399, 2017.

[22] Show in Context CrossRef Check for this item at the UTEP Library Google Scholar

[23] S. Barocas, M. Hardt and A. Narayanan, "Fairness and machine learning", 2019, [online] Available: http://www.fairmlbook.org.

[24] "Draft ethics guidelines for trustworthy AI" in , Brussels, Belgium:European Commission, 2018.

[25] F. Tramèr et al., "Fairtest: Discovering unwarranted associations in data-driven applications", Proc. IEEE Eur. Symp. Secur. Privacy, pp. 401-416, 2017.

[26] M. Feldman, S. A. Friedler, J. Moeller, C. Scheidegger and S. Venkatasubramanian, "Certifying and removing disparate impact", Proc. 21th ACM SIGKDD Int. Conf. Knowl. Discovery Data Mining, pp. 259-268, 2015.

[27] O. Bastani, X. Zhang and A. Solar-Lezama, "Probabilistic verification of fairness properties via concentration", Proc. ACM Program. Languages, pp. 118:1-118:27, 2019.

[28] C. Dwork, M. Hardt, T. Pitassi, O. Reingold and R. S. Zemel, "Fairness through awareness", Proc. Innovations Theor. Comput. Sci., pp. 214-226, 2012.

[29] S. Garg, V. Perot, N. Limtiaco, A. Taly, E. H. Chi and A. Beutel, "Counterfactual fairness in text classification through robustness", Pro. AAAI/ACM Conf. AI Ethics Soc., pp. 219-226, 2019.

[30] P. S. Thomas, B. C. da Silva, A. G. Barto, S. Giguere, Y. Brun and E. Brunskill, "Preventing undesirable behavior of intelligent machines", Science, vol. 366, no. 6468, pp. 999-1004, 2019.

[31] S. Galhotra, Y. Brun and A. Meliou, "Fairness testing: Testing software for discrimination", Proc. 11th Joint Meeting Foundations Softw. Eng., pp. 498-510, 2017.

[32] S. Udeshi, P. Arora and S. Chattopadhyay, "Automated directed fairness testing", Proc. 33rd ACM/IEEE Int. Conf. Automated Softw. Eng., pp. 98-108, 2018.

[33] A. Aggarwal, P. Lohia, S. Nagar, K. Dey and D. Saha, "Black box fairness testing of machine learning models", Proc. ACM Joint Meeting Eur. Softw. Eng. Conf. Symp. Foundations Softw. Eng., pp. 625-635, 2019.

[34] M. T. Ribeiro, S. Singh and C. Guestrin, "", Proc. 22nd ACM SIGKDD Int. Conf. Knowl. Discovery Data Mining, pp. 1135-1144, 2016.

[35] Goodfellow, Ian J., Jonathon Shlens and Christian Szegedy. “Explaining and Harnessing Adversarial Examples.” *CoRR* abs/1412.6572 (2014): n. page